\documentclass[letterpaper]{article}
\usepackage{aaai2026}
\copyrighttext{\hspace*{1.8em}Preprint.}
\usepackage{times}
\usepackage{helvet}
\usepackage{courier}
\usepackage[hyphens]{url}
\usepackage{graphicx}
\usepackage{natbib}
\usepackage{caption}
\usepackage[utf8]{inputenc}
\usepackage[T1]{fontenc}
\usepackage{booktabs}
\usepackage{amsfonts}
\usepackage{nicefrac}
\usepackage{microtype}
\usepackage{xcolor}
\usepackage{tabularx}
\usepackage{adjustbox}
\usepackage{subcaption}
\usepackage{array}

\newcommand{\opensciref}{open-sci-ref-0.02}

\title{open-sci-ref-0.02: Open Reference Baselines for Reproducible Language Model and Dataset Comparison}

\author{
    Marianna Nezhurina\textsuperscript{\rm 1,2,3,4}\equalcontrib{} Harsh Raj\textsuperscript{\rm 1,2,3,4}\equalcontrib{} Victor May\textsuperscript{\rm 9}\equalcontrib{} Joerg Franke\textsuperscript{\rm 1,3,4,5,8} Taishi Nakamura\textsuperscript{\rm 3,6} \\ Timur Carstensen\textsuperscript{\rm 3,4,5,8}
    Niccolò Ajroldi\textsuperscript{\rm 3,4,5} Ville Komulainen\textsuperscript{\rm 3,4,7} David Salinas\textsuperscript{\rm 3,4,5,8} Jenia Jitsev\textsuperscript{\rm 1,2,3,4}\equalcontrib{}
}
\affiliations{
    \textsuperscript{\rm 1}LAION \quad \textsuperscript{\rm 2}Juelich Supercomputing Center (JSC), Research Center Juelich (FZJ) \quad \textsuperscript{\rm 3}Open-$\Psi$ (Open-Sci) Collective  \\
    \textsuperscript{\rm 4}OpenEuroLLM team \quad \textsuperscript{\rm 5}ELLIS Institute Tübingen \quad \textsuperscript{\rm 6}Institute of Science Tokyo  \\
    \textsuperscript{\rm 7}University of Turku \quad \textsuperscript{\rm 8}University of Freiburg \quad \textsuperscript{\rm 9}Ontocord

}

\begin{document}
\pagestyle{plain}

\maketitle
\thispagestyle{plain}

\begin{abstract}
Reliable progress in language modeling depends on knowing whether a new dataset, architecture, or training recipe is genuinely better than existing alternatives. However, comparisons are rarely reproducible: published models differ in architecture, tokenization, evaluation protocol, and compute budget, and the intermediate artifacts needed to analyze or extend results are seldom released.  We present \texttt{open-sci-ref}, a family of dense transformers trained as open reference baselines at four scales (0.13B--1.7B parameters) on nine main reference datasets, with token budgets up to 1T, using a single tuned and fully documented training procedure.
To facilitate a shared experimentation and measurement infrastructure, we release model weights, intermediate checkpoints, training logs, code, and downstream evaluations. When comparing training procedures, we take into account total compute spent. Placing training procedures on a common compute axis helps identify reference points that are weak or strong for the compute they use, without requiring a full scaling law study for every comparison. We identify Nemotron-CC-HQ as the strongest reference dataset on our evaluation suite: our 1.7B model trained on 1T of its tokens matches SmolLM2-1.7B with roughly 11$\times$ less compute. Shared baselines expose pitfalls of scaling trend comparisons: when a small-scale point is unusually weak, the resulting slope can overstate the scalability of a training procedure. We illustrate this diagnostically with EuroLLM-1.7B, which underperforms our 300B-token FineWeb-Edu baseline despite 13$\times$ more training compute. As an additional case study, the suite supports controlled comparison between permissively licensed and mixed-license datasets: the permissive-first MixtureVitae dataset changes rank substantially with scale. This also illustrates why data ranking should be evaluated through multi-scale reference trajectories rather than single proxy runs. All artifacts are publicly released.\footnote{Model weights and intermediate checkpoints are available at \url{https://huggingface.co/collections/open-sci/open-sci-ref-002}. Code for reproducing training, evaluation and raw experiments data is available at \url{https://github.com/LAION-AI/open-sci-ref-0.02}.}

\end{abstract}

\section{Introduction}
\label{sec:intro}

\begin{figure*}[tbh]
    \centering
    \includegraphics[width=.9\textwidth]{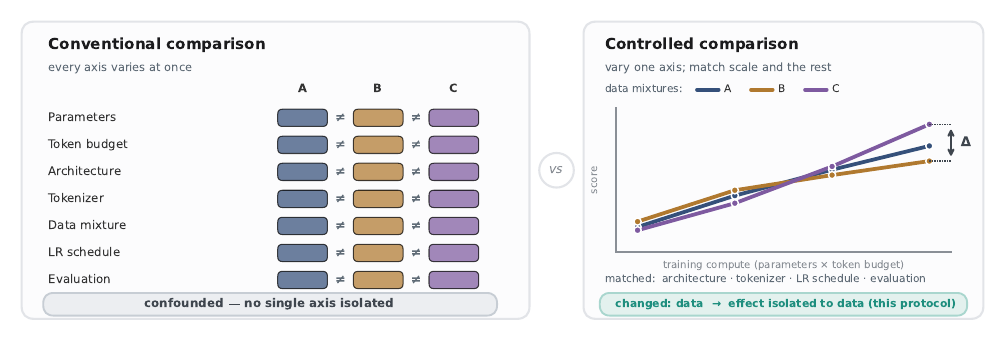}
 \caption{\textbf{Why shared reference baselines are needed.} \textit{Left}:
published models differ on every axis at once, so an observed gap cannot be
attributed to any single cause. \textit{Right}: varying one factor (here the
pretraining data) while matching the rest and placing runs on a common
compute axis isolates its effect under a shared protocol. Rank order may not be
preserved across scales.}
    \label{fig:hero}
\end{figure*}

\begin{table*}[htbp]
\centering
\begin{adjustbox}{width=\textwidth}
\begin{tabular}{@{}lcccccccccc@{}}
\toprule
Model & Dataset & Tokens & \begin{tabular}[c]{@{}c@{}}Params\\(B)\end{tabular} & \begin{tabular}[c]{@{}c@{}}Compute\\(FLOPs)\end{tabular} & Avg & \begin{tabular}[c]{@{}c@{}}ARC-C\\{[}10{]}\end{tabular} & \begin{tabular}[c]{@{}c@{}}ARC-E\\{[}10{]}\end{tabular} & \begin{tabular}[c]{@{}c@{}}HellaSwag\\{[}10{]}\end{tabular} & \begin{tabular}[c]{@{}c@{}}MMLU\\{[}5{]}\end{tabular} & \begin{tabular}[c]{@{}c@{}}COPA\\{[}0{]}\end{tabular} \\
\midrule
\texttt{HF-ref}-1.7B & C4 & 350B & 1.7 & $3.57 \cdot 10^{21}$ & 0.506 & 0.32 & 0.66 & 0.64 & 0.25 & 0.74 \\
\texttt{open-sci-ref}-1.7B & C4 & 300B & 1.7 & $3.06 \cdot 10^{21}$ & 0.548 & 0.34 & 0.67 & 0.68 & 0.26 & 0.81 \\
\texttt{HF-ref}-1.7B & FineWeb-Edu & 350B & 1.7 & $3.57 \cdot 10^{21}$ & 0.541 & 0.46 & 0.77 & 0.62 & 0.25 & 0.78 \\
\texttt{open-sci-ref}-1.7B & FineWeb-Edu & 300B & 1.7 & $3.06 \cdot 10^{21}$ & 0.567 & 0.48 & 0.77 & 0.67 & 0.26 & 0.82 \\
\bottomrule
\end{tabular}
\end{adjustbox}
\caption{Comparing HuggingFace (HF) and \texttt{open-sci-ref} reference baselines on C4 and FineWeb-Edu-1.4T using standardized evaluations. [n] denotes the few-shot example counts. \texttt{open-sci-ref} (300B tokens) matches or outperforms the results of HF (350B tokens) reference baseline. The ``Avg'' column reports the 11-task average.}
\label{tab:sanity_check}
\end{table*}

\begin{table*}[ht]
\centering
\footnotesize
\begin{tabular}{l c c c c c c}
\hline
\shortstack{Params (B)\\(Non-Emb + Emb)} & Layers & Hidden & Heads & \shortstack{FFN\\Hidden} & \shortstack{GPU Memory\\Usage} & \shortstack{FLOPs/token\\(6N estimate)} \\
\hline
0.1 + 0.03 = 0.13 & 22 & 512  & 8  & 2256 & 0.89GB  & $7.8 \cdot 10^{8}$  \\
0.35 + 0.05 = 0.40 & 22 & 1024 & 16 & 3840 & 2.32GB  & $2.4 \cdot 10^{9}$  \\
1.21 + 0.10 = 1.31 & 24 & 2048 & 32 & 5440 & 7.54GB & $7.9 \cdot 10^{9}$  \\
1.61 + 0.10 = 1.71 & 24 & 2048 & 32 & 8192 & 9.88GB & $1.0 \cdot 10^{10}$ \\
\hline
\end{tabular}
\caption{\texttt{open-sci-ref} model architecture and scales.}
\label{tab:model_architecture}
\end{table*}

When a new pretraining dataset, training procedure, or architecture is declared an improvement over an existing reference, the evidence is a comparison against other published models. Yet published models differ on nearly every axis at once: parameter count, token budget, architecture, tokenizer, data mixture, learning rate schedule. Therefore, aligned comparison is hard and an observed gain is hard to attribute to any single cause (Fig.~\ref{fig:hero}). Evaluation setup compounds the problem: the same benchmark scored under different evaluation setups, such as the few-shot number or sampling strategy, yields scores that are not directly comparable across papers.

This problem is especially acute for dataset and scaling studies. A dataset that appears strong at one training scale may not remain strong at another. A smaller scale model that appears competitive by parameter count against a larger one may have used far more total training compute. A scaling trend can also look steep simply because the smallest model in the trend was poorly trained. Without shared reference points, researchers must compare against heterogeneous external models whose training details, intermediate checkpoints, or evaluation settings may be unavailable. This makes it difficult to validate new training pipelines, diagnose failures, estimate scaling behavior, or decide whether an experimental dataset will retain its advantage across scales and is worth scaling further.

What is needed instead are open reference baselines: training and evaluation pipelines resulting in models that are not standalone artifacts comparable only within the same study but are also references that other work can measure against, using the same pipelines for the training and evaluation. Serving that role places two demands on a baseline. First, it must be fully specified and reproducible - dataset and its composition, tokenizer, architecture, optimizer, schedule, compute budget, and evaluation protocol all documented. Especially accounting for total training compute is important, as this helps isolate the effect of interventions across different procedures on the differences in their outcome, while explicitly controlling for differences induced by different compute budget. Second, it must cover a broad span of model and token scales to estimate scaling trends or derive full scaling laws, as looking at difference in scores for one scale can be highly misleading if expecting the same trends to continue at higher unseen scales. Together, rigor for each executed run including intermediate checkpoints beyond only the final one and the sufficient scale span to estimate scaling trends are what let a baseline serve as anchor for robust comparison. Such comparison to further controlled experiments can help identify which interventions improve the training procedure and which do not, enabling guided search for better scalable learning procedures.

\texttt{open-sci-ref} builds on open training suites, controlled data selection testbeds, standardized evaluation frameworks, and compute-aware scaling studies. Its distinct role is to provide reusable trajectories for direct comparison among independently released pretraining corpora under one shared protocol. Section~\ref{sec:related-work} situates this design relative to complementary efforts.

The goal of \texttt{\opensciref{}} is to fill this gap. The comparisons that motivate this work vary on many axes at once - ours proposes a canonical setup and runs all experiments against it. We train a family of dense transformers under a tuned and fully documented procedure, applied unchanged across four model scales (0.13B, 0.4B, 1.3B, and 1.7B parameters), nine reference datasets, and token budgets of 50B, 300B, and 1T. Within each matched model size and token budget setting, observed downstream differences estimate the effect of substituting the pretraining dataset under the shared tokenizer, architecture family, training procedure, and evaluation setup. Across settings, the resulting trajectories show how this protocol-specific comparison changes with scale, Appendix~\ref{app:scope-of-control} details its scope and limitations. We release the full trajectory behind each run - final weights, intermediate checkpoints, training logs, code, and standardized evaluations - turning the reference baselines into shared measurement infrastructure: any new dataset, recipe, or model can be aligned with them on a common compute axis and read against known reference points.

Among the conventional web-oriented reference datasets, rankings are broadly stable across model scales and throughout training, while MixtureVitae provides an important exception: its relative rank improves substantially with scale. The procedure is also strong in absolute terms: our 1.7B model trained on 1T tokens of Nemotron-CC-HQ matches SmolLM2-1.7B~\cite{allal2025smollm2} on an 11-task evaluation suite while using roughly 11$\times$ less compute. Aligning external models on the same compute axis yields  a further lesson: it exposes cases where steep scaling trends are artifacts of weak small-scale models.

The same reference suite also enables a controlled comparison between
permissively licensed datasets and mixed- or unspecified-license references.
In this release we extend the permissive set beyond CommonCorpus-en by adding
Comma-0.1 and MixtureVitae. This comparison is not meant to settle the general
legal or data curation question. Instead, it asks what happens when datasets
with different sourcing and mixture choices are trained under the same
architecture, tokenizer, optimization schedule, compute budget, and evaluation
protocol. The resulting trajectories show that permissive sourcing alone is
not predictive of performance: CommonCorpus-en and Comma-0.1 trail the
strongest curated references, while MixtureVitae changes rank substantially
with scale.

To summarize, our contributions are:
\begin{itemize}
    \item An open, reproducible suite of reference training runs spanning nine main dataset comparison corpora, four model scales, and three token budgets, released with all intermediate artifacts.
    \item A simple methodology for apples-to-apples comparison of new training procedures and datasets.
    \item A controlled comparison of permissively licensed and mixed-licensed reference datasets under one fixed training and evaluation procedure.
\end{itemize}

\section{Related Work}
\label{sec:related-work}

\paragraph{Open training suites and trajectories.}
Open model suites increasingly release artifacts beyond final weights. Pythia provides dense checkpoints across model scales while holding the training data and order fixed
~\cite{biderman2023}, and PolyPythias extends this design across random seeds to study training stability~\cite{vanderwal2025}. OLMo~\cite{groeneveld-etal-2024-olmo,olmo20242} releases data, code,
checkpoints, and evaluation artifacts for an inspectable model development pipeline.
These projects establish the value of open training trajectories, but primarily organize their experiments around
model scale, training stage, or random seed within one principal data pipeline. \texttt{open-sci-ref} instead makes the pretraining dataset a primary experimental axis, applying
one documented protocol across nine independently
constructed corpora and releasing the resulting trajectories as reusable reference points.

\paragraph{Controlled comparison of pretraining data.}
DataComp-LM (DCLM) provides a standardized testbed for comparing data curation strategies over a shared Common Crawl candidate pool, using fixed training and evaluation procedures~\cite{li2025}. DataDecide is closest to our objective: it
compares 25 pretraining corpora using models up to 1B parameters and budgets up to 100B tokens, with the goal of determining whether inexpensive experiments predict larger-scale data choices~\cite{magnusson2025datadecide}.

\texttt{open-sci-ref} is complementary to these efforts. Rather than evaluating curation methods over one candidate pool or optimizing predictions from small proxy runs, we compare distinct released corpora under a common protocol.
We provide trajectories across four model scales at the 300B-token budget, together with selected 1.7B-parameter extensions to 1T tokens. The objective is to supply transparent control runs against which new datasets, training procedures, and scale-dependent predictions can be
tested directly.

\paragraph{Evaluation and compute-aware comparison.}
Frameworks such as the LM Evaluation Harness and HELM standardize task definitions, prompting conditions, and metrics~\cite{eval-harness, liang2023holistic}, while scaling law research relates performance to model size,
training data, and compute~\cite{Kaplan2020, hoffmann2022, hagele2024scaling}.
\texttt{open-sci-ref} connects these traditions. It is neither
a new downstream benchmark nor a full compute-optimal scaling study. It combines a standardized evaluation setup with a standardized training protocol and releases the intermediate states of that protocol. This allows new procedures to be compared with matched reference runs, placed on a common estimated pretraining compute axis, and examined throughout training rather than only at their final checkpoints.

\begin{figure*}[t!]
    \centering
    \includegraphics[width=.9\textwidth]{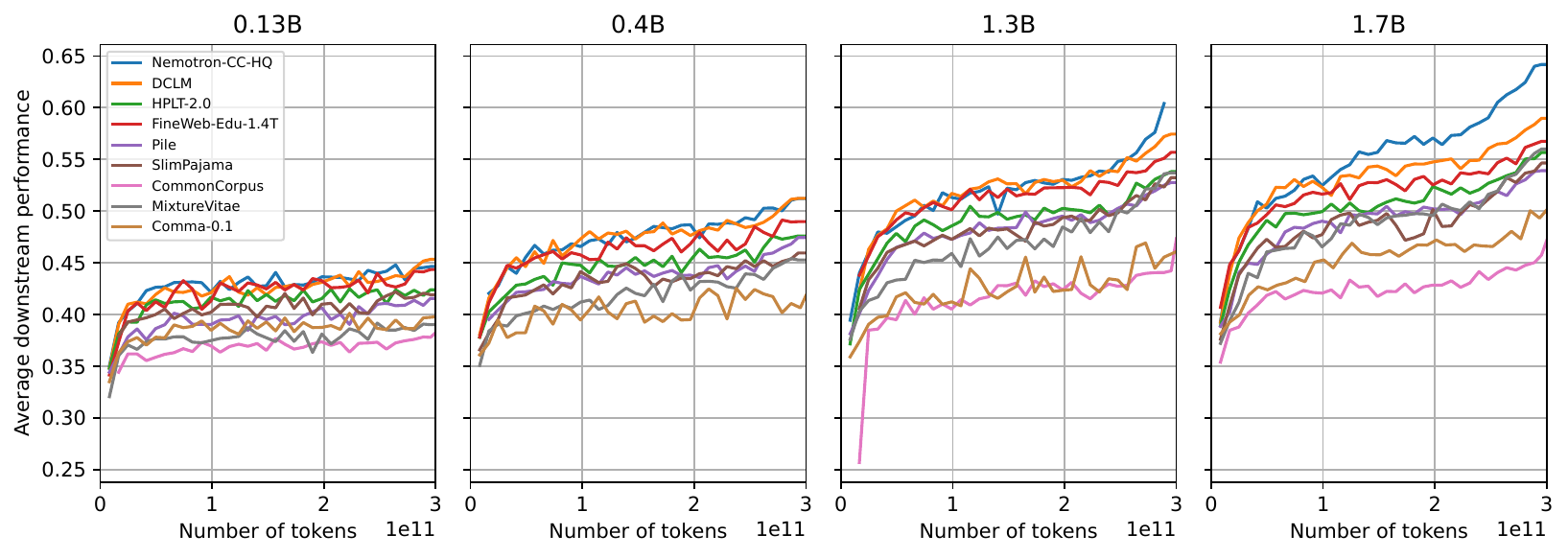}
    \caption{Average downstream performance over training for full reference trajectories at four model scales. Rankings are broadly stable among conventional datasets, with gaps widening at larger scales. MixtureVitae is the main exception: it begins near the bottom at 0.13B but rises to the HPLT/FineWeb-Edu group at 1.7B, illustrating why single-scale dataset rankings can be misleading.}
    \label{fig:reference300b}
\end{figure*}

\section{Methods \& Experiment Setup}
\label{sec:methods_experiment}

\subsection{Datasets, tokenization, and training details}

For the main dataset comparison suite we consider nine datasets. Six are established references with mixed or unspecified licensing:
Pile~\cite{gao2020pile}, SlimPajama~\cite{cerebras2023slimpajama},
FineWeb-Edu-1.4T (v1.0.0)~\cite{penedo2024}, DCLM-baseline~\cite{li2025},
Nemotron-CC-HQ~\cite{su-etal-2025-nemotron}, and HPLT-2.0~\cite{burchell2025}.
Three are permissively licensed: CommonCorpus-en~\cite{langlais2026common},
Comma-0.1, derived from the Common Pile~\cite{kandpal2025commonpilev018tb},
and MixtureVitae~\cite{nguyen2026mixturevitae}. For HPLT-2.0 and CommonCorpus-en we train on the English subset, so that all nine are compared
with language held approximately fixed. C4~\cite{raffel2020exploring} is not among the nine. We use it only as an auxiliary corpus for recipe tuning, training procedure validation, and as an additional reference line in the compute axis plot.

The previous iteration \texttt{open-sci-ref}-0.01~\cite{nezhurina2025opensciref}
included CommonCorpus as its only permissive pool. The two additions make a
direct permissive versus mixed-license comparison possible under one fixed procedure, and they differ in construction. Comma-0.1 follows a conventional design, drawing exclusively on public-domain and openly licensed human-written text. MixtureVitae combines permissive text with low-risk additions and front-loads math, code, reasoning, and instruction data (including a large synthetic subset) into the single pre-training stage, reducing the share of web text
that dominates conventional mixtures. It therefore tests whether the standard web-dominated recipe is necessary for language understanding performance.

We tokenize datasets using the GPT-NeoX-20B tokenizer~\cite{black-etal-2022-gpt},
resulting in a vocabulary size of 50,304. For the learning rate we use a
constant schedule with linear cooldown (warmup-stable-decay, WSD)~\cite{hu2024minicpm_colm,hagele2024scaling}, which unlike cosine decay
does not require the token budget to be fixed in advance (Appendix~\ref{app:hparams}). For all datasets, we train reference dense transformer models at 0.13B, 0.4B, 1.3B, and 1.7B parameters using
Megatron-LM~\cite{megatron-lm}. We describe the architecture in Section~\ref{sec:architecture} and the training recipe in Section~\ref{sec:appHPO}.
Unless stated otherwise, the reference runs use a context length of 4096
tokens with a RoPE base of 100k.

\subsection{Model architecture}
\label{sec:architecture}
We follow a custom design starting from the
Llama~\citep{touvron2023llamaopenefficientfoundation} architecture. Different from standard Llama, we include biases in all linear layers (QKV projections, attention, and FFN), as removing them in ablation experiments at the 1.3B, 50BT
scale resulted in performance deterioration, and we add QK-Norm~\citep{henry2020qknorm,wortsmansmall} for improved training stability and better performance, following recent open models such as OLMo~2~\citep{olmo20242} and Qwen3~\citep{yang2025qwen3}. We use
SwiGLU~\citep{shazeer2020glu} and tied embedding weights across all model scales. Hereafter, we refer to this model design as \texttt{open-sci-ref}. The exact configurations are given in Tab.~\ref{tab:model_architecture}.

\begin{figure*}[ht!]
    \centering
    \includegraphics[width=.9\textwidth]{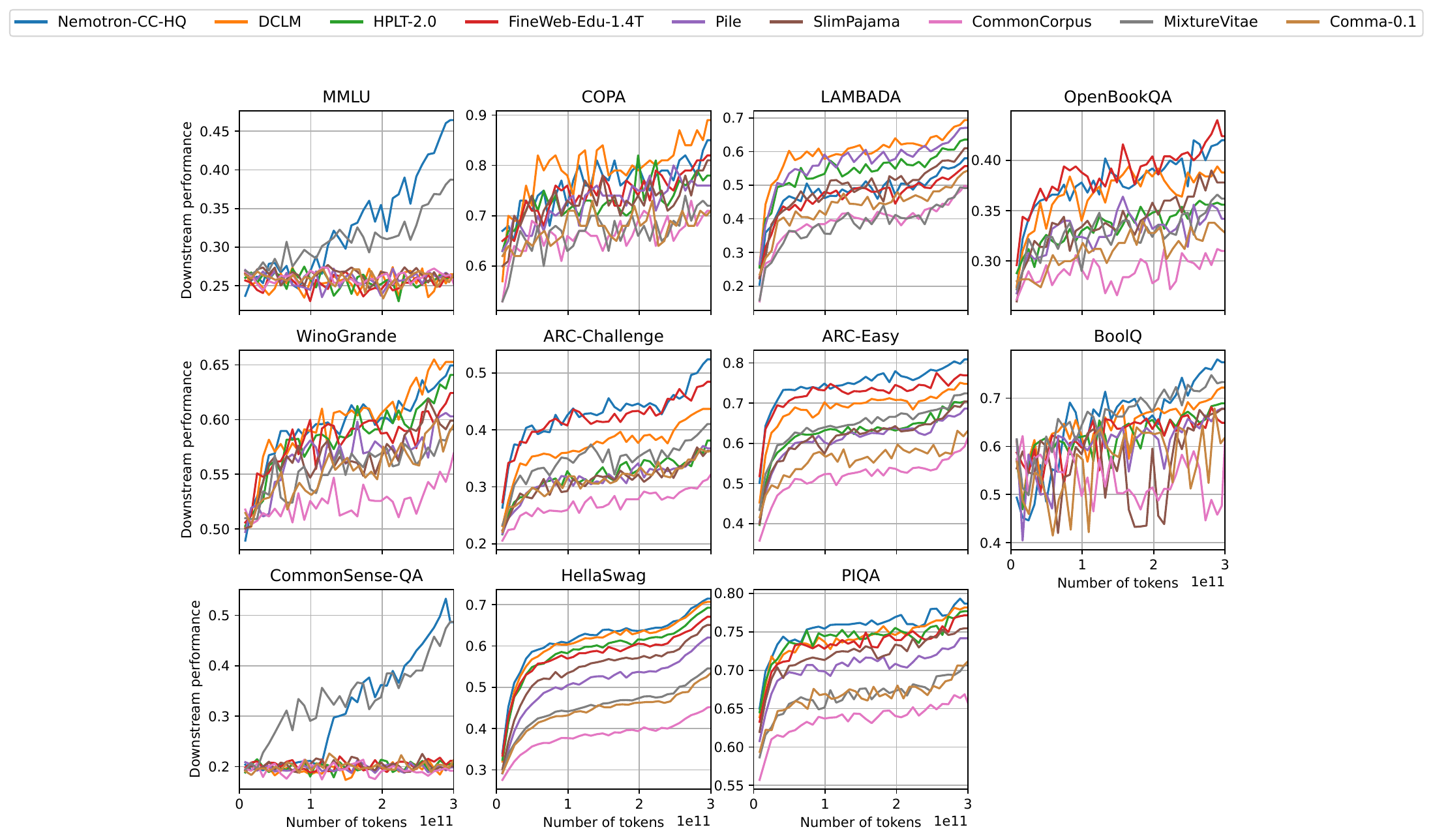}
    \caption{Downstream performance across 11 tasks of \texttt{open-sci-ref} models with 1.7B parameters trained on 300B tokens over different datasets.  While some tasks yield clear dataset rankings (e.g., ARC, HellaSwag, LAMBADA), others provide insufficient signal for meaningful dataset comparison.}
    \label{fig:reference_17b_300b}
\end{figure*}

\subsection{Hyperparameter tuning}
\label{sec:appHPO}

We tune the training recipe at the 1.3B scale using 50B-token runs on C4 and FineWeb-Edu, selecting configurations by validation loss. We first tune standard cosine learning rate schedules, then switch to a WSD schedule and tune its learning rate until it matches or improves on the best cosine validation loss. The final reference runs use the WSD schedule with weight decay 0.05, a global batch size of approximately 4M tokens, and a cooldown lasting 20\% of the total training duration. For token budgets of 300B and above, we use a longer warmup of 25,000 iterations. Full tuning and schedule details are reported in Appendix~\ref{app:hparams}. 

\subsection{Training recipe validation and downstream evaluation}
\label{sec:recipe}
To validate the tuned training procedure of our reference models, we perform a comparison with established baselines on the open datasets C4 and FineWeb-Edu, using open-source HuggingFace (HF) models with 1.7B parameters, trained on 350B tokens (denoted as \texttt{HF-ref}). We evaluate both \texttt{HF-ref} and \texttt{open-sci-ref} on standardized benchmarks (Tab. \ref{tab:sanity_check}). Since \texttt{open-sci-ref} matches or outperforms \texttt{HF-ref} - despite the latter being trained on 50B more tokens - we conclude that our training procedure is sound and well tuned at the 1.7B model scale.

\textbf{Evaluation.} Having established the soundness of our training procedure, we execute training of reference baselines and evaluate every intermediate checkpoint using LM-eval-harness~\cite{eval-harness}. We evaluate on COPA~\cite{copa}, OpenBookQA~\cite{mihaylov-etal-2018-suit}, LAMBADA~\cite{paperno2016lambada} and WinoGrande~\cite{sakaguchi2019winogrande} in the 0-shot setting, MMLU~\cite{hendrycks2021measuring} in 5-shot, and CommonSenseQA~\cite{talmor2019commonsenseqa}, PIQA~\cite{bisk2020piqa}, ARC-Challenge and ARC-Easy~\cite{clark2018thinkarx}, HellaSwag~\cite{zellers2019hellaswag} and BoolQ~\cite{clark2019boolq} with 10-shot, resulting in 11 evaluations in total. We report length-normalized
accuracy (lm-eval's \texttt{acc\_norm}) for ARC-Challenge, ARC-Easy, HellaSwag,
OpenBookQA, and PIQA, and accuracy (\texttt{acc}) for the remaining six tasks.

\textbf{Control experiments.} We perform additional experiments, testing the effect of different context lengths (4096, 8192, 16384 tokens) on the 1.7B model trained for 100B tokens (Suppl. Sec. \ref{subsec:further_reference_baselines}).

\section{Results}\label{sec:results}

\begin{figure*}[htb]
    \centering
    \includegraphics[width=.7\textwidth]{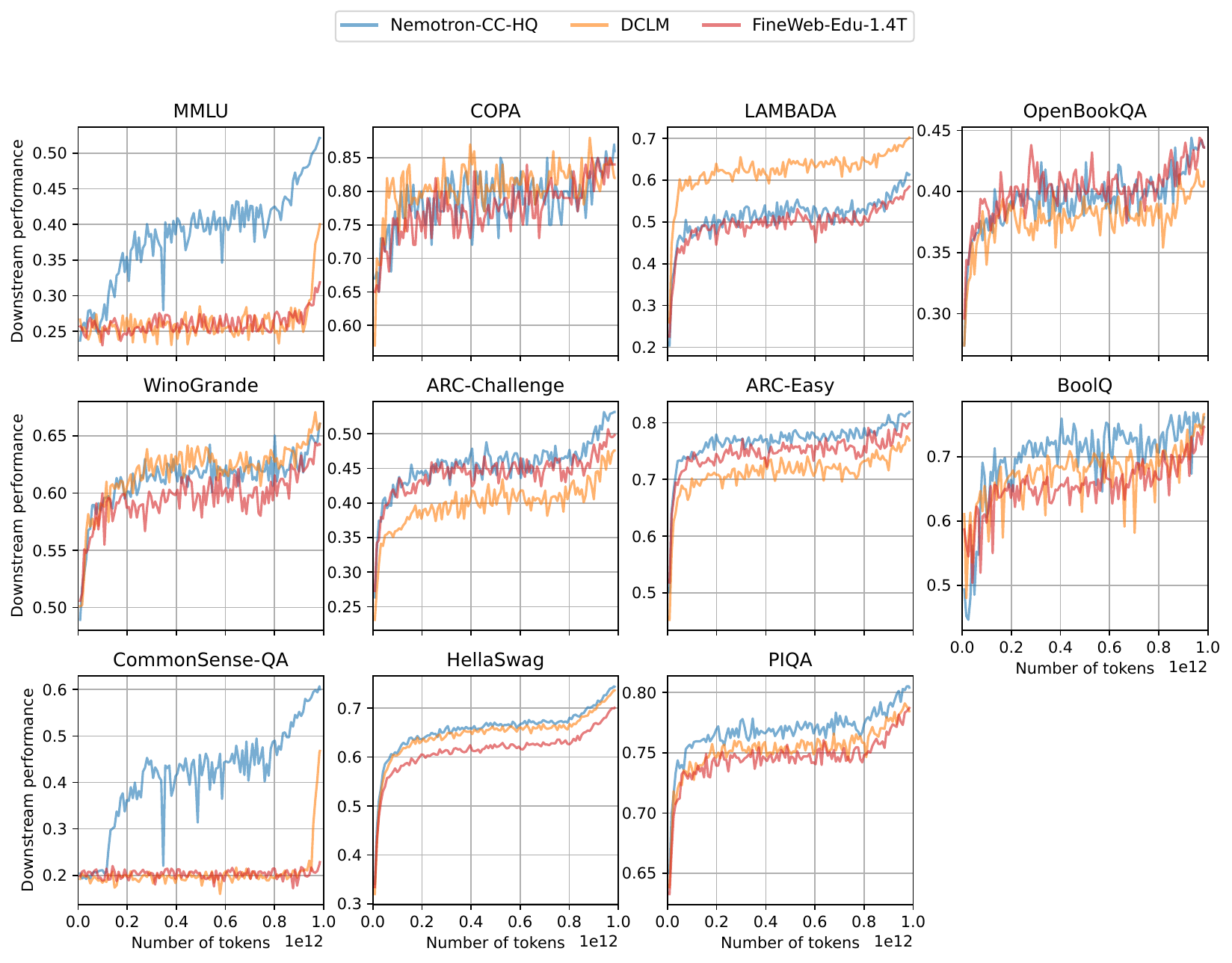}
    \caption{Downstream performance on 11 tasks for
\texttt{open-sci-ref} 1.7B models trained for 1T tokens on the top three
reference datasets. Nemotron-CC-HQ has the strongest average endpoint,
while DCLM is strongest on LAMBADA and competitive on several continuation-style tasks.}
    \label{fig:reference_17b_1t}
\end{figure*}

\subsection{Dataset comparison across scales}
\label{sec:dataset-comparison}

Figure~\ref{fig:reference300b} reports the 11-task average over
training for four model sizes at the 300B-token budget. Among the conventional
web-oriented reference datasets, rankings are broadly stable across model
scale: Nemotron-CC-HQ and DCLM form the strongest group, followed by
FineWeb-Edu and HPLT-2.0, with Pile and SlimPajama below them.
The gaps also widen at larger model scales, making the 1.7B runs more
informative than the smallest proxy setting. Appendix~\ref{app:aggregate-rank-robustness}
quantifies the stability of these rankings across scales and their sensitivity to
the composition of the evaluation suite.

The stability is not universal. MixtureVitae changes rank substantially with
scale, moving from near the bottom at 0.13B to the HPLT/FineWeb-Edu range at
1.7B.

Figure~\ref{fig:reference_17b_300b} shows that the 11-task average hides
different endpoint profiles. ARC, LAMBADA, HellaSwag, and PIQA separate the
curated web-oriented datasets more clearly, while MMLU provides limited signal
for most datasets at this scale. The two datasets with the strongest MMLU
signal, Nemotron-CC-HQ and MixtureVitae, are also the two with substantial
instruction-like or reasoning-oriented material in the pretraining mixture.
This suggests that MMLU at 1.7B is not simply a generic dataset quality
measure: it is especially sensitive to the presence of instruction-like
formats and reasoning content. Table~\ref{tab:standard-endpoint-results} reports the corresponding
single-benchmark endpoint values in the main text. This makes the reference
points directly comparable to previous work while preserving the distinction
between controlled \texttt{open-sci-ref} runs and external models.

Figure~\ref{fig:reference_17b_1t} extends the strongest 1.7B references
to 1T tokens. Nemotron-CC-HQ provides the strongest average endpoint, DCLM
remains especially strong on continuation-style tasks such as LAMBADA, and
FineWeb-Edu remains a strong curated-web reference but below the top two on
the 11-task average. These 1T runs provide the high-performing anchors for
the compute axis comparison in the next section.

\begin{table*}[t]
\centering
\scriptsize
\setlength{\tabcolsep}{3pt}
\begin{tabular}{@{}lrrrr rrrrrrrrrrr@{}}
\hline
Model / dataset & Tok. & Par. & FLOPs & Avg.
& COPA & LAMB. & OBQA & Wino. & MMLU
& ARC-C & ARC-E & BoolQ & CSQA & HSwag & PIQA \\
 & (T) & (B) & ($10^{21}$) & & & & & & & & & & & & \\
\hline
\multicolumn{16}{@{}l}{\textit{Controlled \texttt{open-sci-ref-0.02} endpoints (all \texttt{open-sci-ref-1.7B})}} \\
Nemotron-CC-HQ  & 1.0 & 1.7 & 10.2 & \textbf{.67}
& \textbf{.86} & .61 & \textbf{.44} & \textbf{.66} & \textbf{.52}
& \textbf{.53} & \textbf{.82} & .76 & \textbf{.60} & \textbf{.74} & \textbf{.80} \\
DCLM            & 1.0 & 1.7 & 10.2 & .64
& .82 & \textbf{.70} & .41 & \textbf{.66} & .40
& .48 & .77 & \textbf{.77} & .47 & .74 & .78 \\
FineWeb-Edu     & 1.0 & 1.7 & 10.2 & .60
& .84 & .58 & \textbf{.44} & .64 & .32
& .50 & .80 & .75 & .23 & .70 & .79 \\
FineWeb-Edu     & 0.3 & 1.7 & 3.06 & .57
& .82 & .56 & .42 & .62 & .26
& .48 & .77 & .65 & .21 & .67 & .77 \\
MixtureVitae    & 0.3 & 1.7 & 3.06 & .56
& .72 & .49 & .36 & .59 & .39
& .41 & .72 & .73 & .49 & .55 & .71 \\
Comma-0.1       & 0.3 & 1.7 & 3.06 & .50
& .71 & .54 & .33 & .60 & .27
& .36 & .63 & .62 & .21 & .53 & .71 \\
CommonCorpus-en & 0.3 & 1.7 & 3.06 & .47
& .71 & .49 & .31 & .56 & .25
& .32 & .61 & .62 & .19 & .45 & .66 \\
\hline
\multicolumn{16}{@{}l}{\textit{External models, not controlled comparisons}} \\
Gemma-2-2B~\cite{team2024gemma} & 2.0 & 2.6 & 31.2 & .68
& .88 & .70 & .37 & .69 & .53
& .52 & .82 & .80 & .65 & .74 & .80 \\
Qwen2.5-1.5B~\cite{qwen2025qwen25technicalreport} & 18.0 & 1.5 & 162 & .67
& .83 & .62 & .36 & .63 & .61
& .52 & .81 & .78 & .76 & .68 & .77 \\
DCLM-1B (DCLM)            & 4.0 & 1.4 & 33.6 & .66
& .90 & .67 & .43 & .68 & .47
& .48 & .78 & .75 & .62 & .74 & .79 \\
SmolLM2-1.7B (SmolLM2)    & 11.0 & 1.7 & 112 & .66
& .82 & .67 & .38 & .66 & .50
& .52 & .80 & .75 & .60 & .73 & .78 \\
HF-ref-1.7B (FineWeb-Edu) & 0.35 & 1.7 & 3.57 & .54
& .78 & .50 & .37 & .58 & .25
& .46 & .77 & .66 & .19 & .62 & .75 \\
EuroLLM-1.7B              & 4.0 & 1.7 & 40.8 & .52
& .74 & .53 & .33 & .59 & .27
& .39 & .73 & .61 & .19 & .60 & .74 \\
\hline
\end{tabular}
\caption{
Per-benchmark endpoint results for selected \texttt{open-sci-ref-0.02}
1.7B reference baselines and external models. All rows in the
upper block are \texttt{open-sci-ref-1.7B} runs trained on the listed
dataset, where architecture, tokenizer, training recipe, evaluation
protocol, and compute accounting are fixed. \textbf{Bold} marks the best
score per evaluation column within this controlled block, determined from
unrounded scores (ties bolded jointly). The
upper block covers the three 1T endpoints, the 300B FineWeb-Edu reference,
and the three permissively licensed datasets.
Figure~\ref{fig:reference300b} shows all nine datasets at 300B. The
lower block reports external models for orientation, with their training
dataset in parentheses where known. These rows are not controlled
comparisons because they differ in data, tokenizer, architecture details,
and training recipe, and are therefore not included in the bolding.
Training tokens are given in trillions and training compute in units of
$10^{21}$ FLOPs, estimated as $6ND$ from the parameter count $N$ and the
training token count $D$. Scores are accuracies reported without the leading zero. For the
\texttt{open-sci-ref} rows, ARC-C, ARC-E, HSwag, OBQA and PIQA use
length-normalized accuracy (Section~\ref{sec:recipe}).
Avg. is the unweighted mean over the 11 listed evaluations, including
CommonSenseQA. LAMB. denotes LAMBADA, OBQA denotes OpenBookQA, Wino.
denotes WinoGrande, CSQA denotes CommonSenseQA, and HSwag denotes
HellaSwag.
}
\label{tab:standard-endpoint-results}
\end{table*}

\begin{figure*}[htb]
    \centering
    \includegraphics[width=0.75\textwidth]{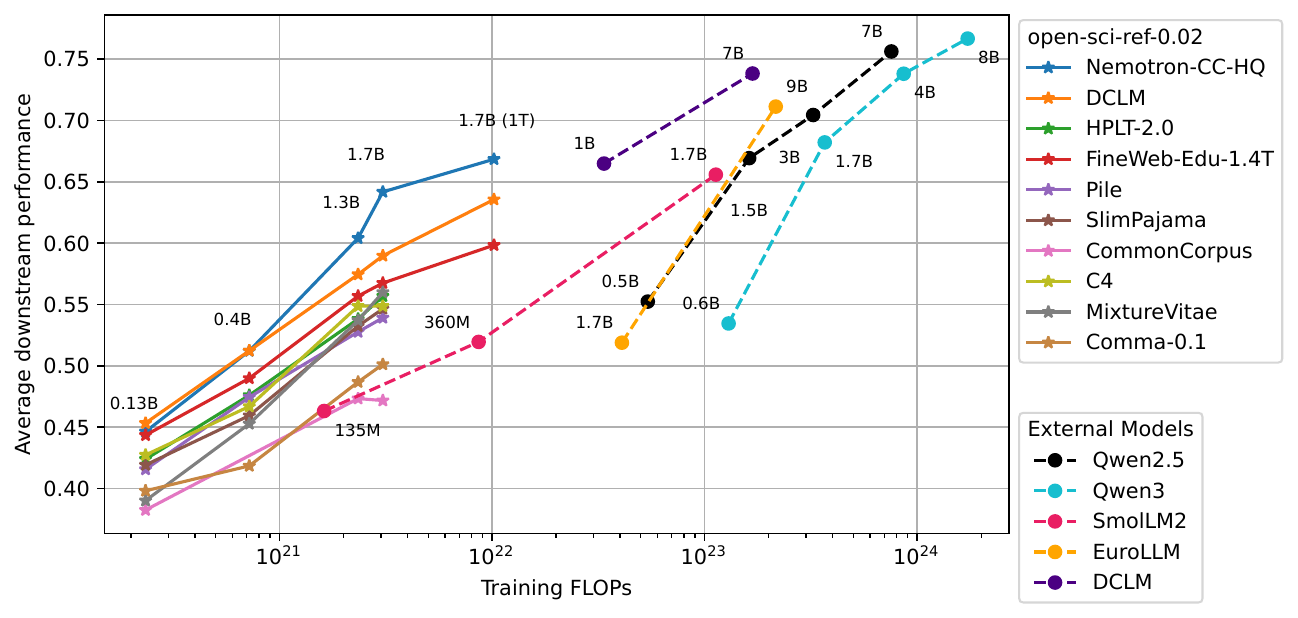}
    \caption{
Scaling trends for \texttt{open-sci-ref-0.02} on a common estimated
training compute axis. Solid lines show controlled \texttt{open-sci-ref}
runs, where architecture, tokenizer, training recipe, evaluation protocol,
and compute accounting are fixed. Dashed lines show selected external
model families: these are not controlled comparisons
because the models differ in data, tokenizer, architecture details, and
training recipe.
}
    \label{fig:scaling_compute}
\end{figure*}

\subsection{Permissive and mixed-license reference datasets}
\label{sec:permissive-comparison}
The three permissively licensed pools occupy distinct positions in the
ranking. CommonCorpus-en and Comma-0.1 remain below every curated mixed- or
unspecified-license reference at every scale, and which of the two is weaker
changes with scale. MixtureVitae is the main exception
to the otherwise broadly stable rank ordering (Sec.~\ref{sec:dataset-comparison}), and
it is the only one of the three whose mixture departs from the conventional
web-dominated recipe.

Table~\ref{tab:standard-endpoint-results} shows that MixtureVitae's
1.7B/300B average is close to FineWeb-Edu's, but the gap it closes comes
almost entirely from MMLU (0.39 against 0.26) and CommonSenseQA (0.49 against
0.21), the two tasks Section~\ref{sec:dataset-comparison} identifies as
sensitive to instruction-like material. Averaged over the remaining nine
tasks, FineWeb-Edu leads 0.64 to 0.59. We therefore treat the permissive
comparison as evidence that licensing status alone is not predictive of
performance. Mixture design, filtering, scale, and task composition all matter.

\subsection{Compute axis comparison}

Figure~\ref{fig:scaling_compute} places the reference trajectories on a common
estimated training compute axis. Solid lines denote controlled
\texttt{open-sci-ref} runs, where architecture, tokenizer, training recipe,
evaluation protocol, and compute accounting are held fixed. Dashed lines show
selected external model families for orientation. These external markers are
not controlled comparisons with \texttt{open-sci-ref}. They are included to
show how public reference trajectories can contextualize published models.

Within the controlled suite, the compute axis view preserves the main dataset
ordering while making the size of the gaps easier to interpret. The Nemotron-CC-HQ 1T endpoint is the strongest reference point in this release,
with DCLM and FineWeb-Edu providing additional high-performing anchors. For orientation, the Nemotron-CC-HQ 1T endpoint lies near the published
SmolLM2-1.7B point while using much less estimated training compute. This comparison is useful context, but not a controlled ablation.

The same plot illustrates a pitfall in reading scaling trends from heterogeneous external models. EuroLLM-1.7B~\cite{martins2024eurollm,martins2025eurollm} lies below several
\texttt{open-sci-ref} reference points trained with less estimated compute,
including the 300B-token FineWeb-Edu run. This does not isolate the effect of multilingual data allocation, since EuroLLM differs in many design choices.
Rather, it shows why common reference curves are useful: a steep trend between two external model sizes can partly reflect a weak small-scale point, not only a more scalable training procedure. Precise scaling law conclusions require
denser measurements near compute-optimal frontiers. The trajectories here are reference comparisons, not full scaling laws.

\subsection{Further insights from multilingual benchmarks}

\begin{figure*}[htbp]
    \centering
    \includegraphics[width=\textwidth]{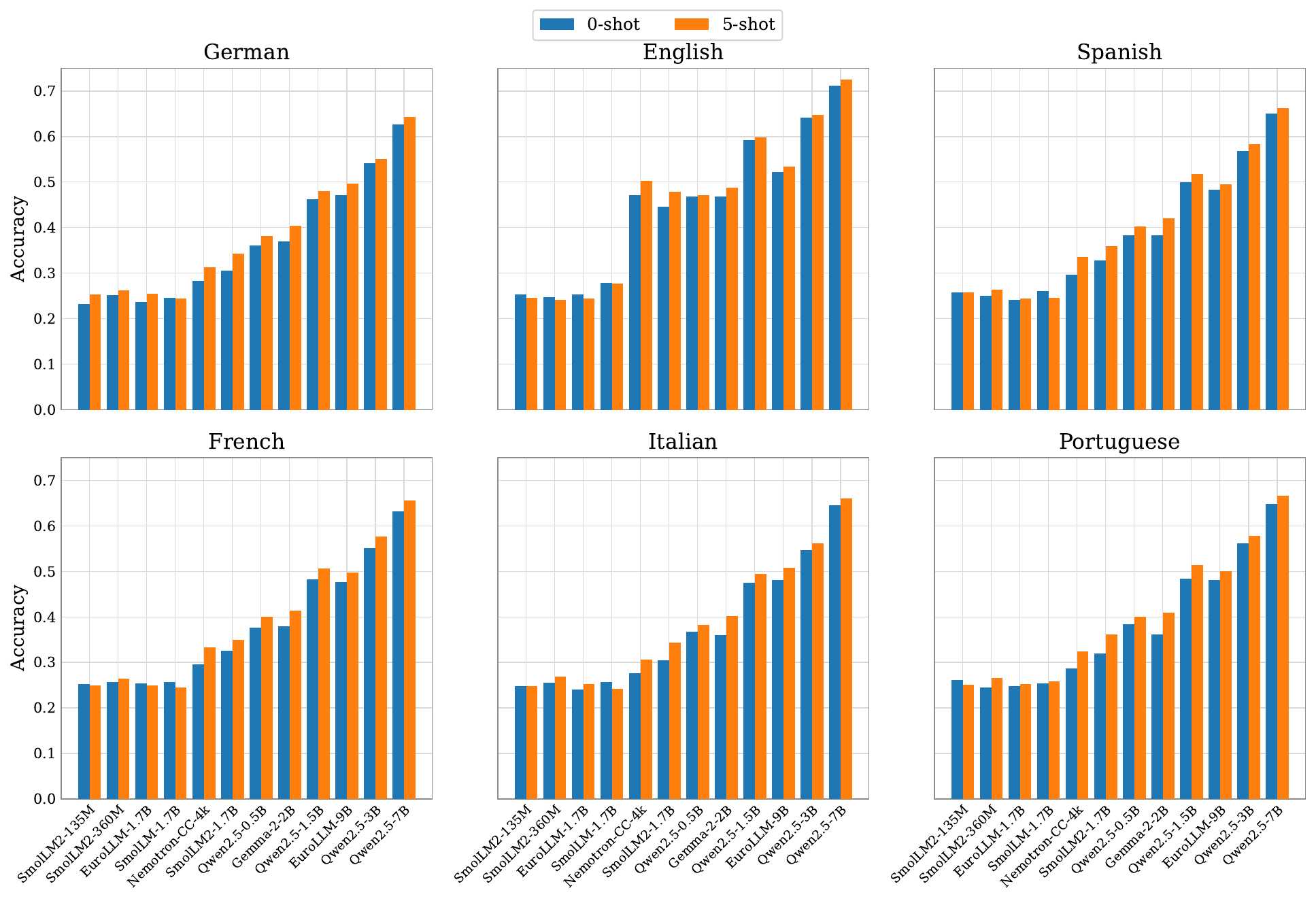}
    \caption{Performance of English-dominant \texttt{open-sci-ref} on MMMLU, which is a version of the MMLU benchmark translated to multiple languages by OpenAI employing human professional translators.}
    \label{fig:mmmlu_comparison}
\end{figure*}

\begin{figure*}[htbp]
    \centering
    \includegraphics[width=\textwidth]{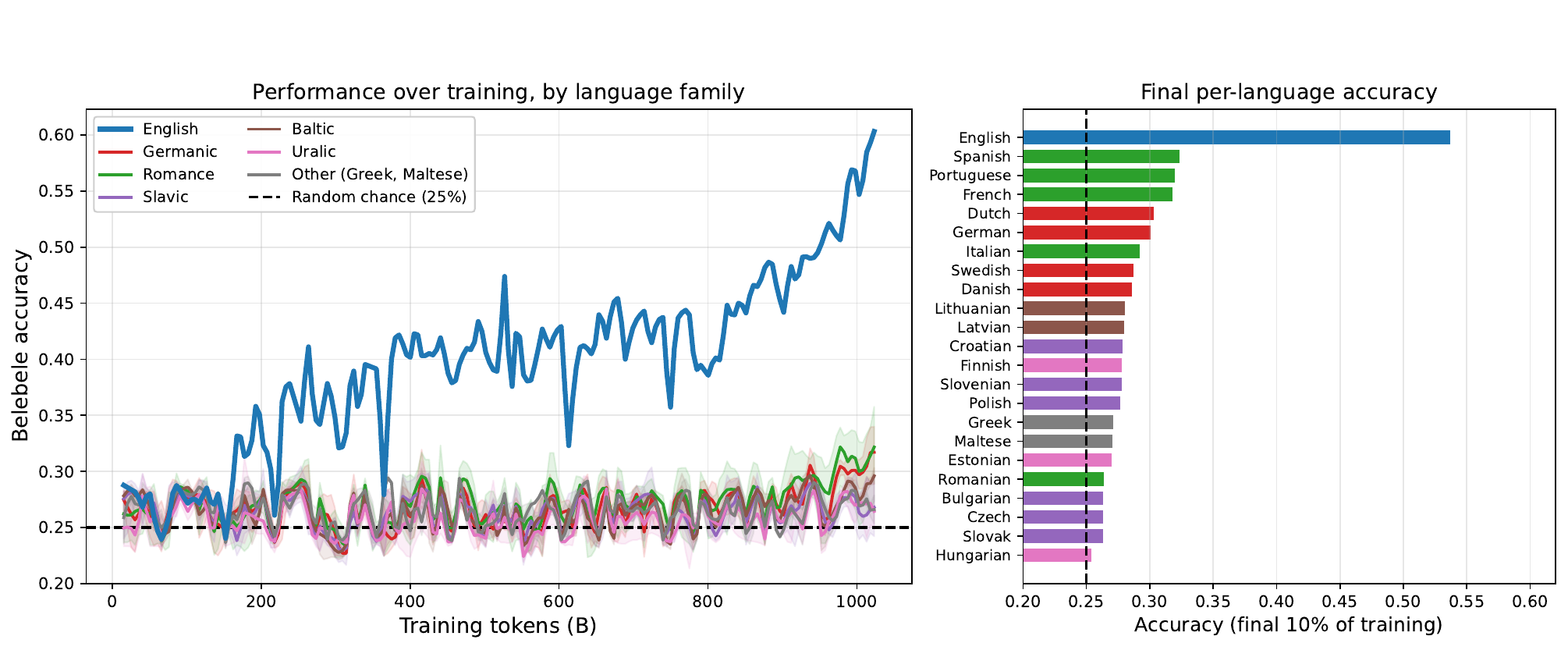}
    \caption{Average performance on Belebele for different languages while training a 1.7B model on Nemotron-CC-HQ.}
    \label{fig:belebele}
\end{figure*}

We also evaluate the Nemotron-CC-HQ 1T model on two multilingual benchmarks:
MMMLU~\cite{openai2024mmmlu}, which contains translated MMLU questions by
OpenAI, and Belebele~\cite{Bandarkar_2024}. These experiments are exploratory and are not part of the
main controlled dataset comparison. Figure~\ref{fig:mmmlu_comparison} shows that the
Nemotron-CC-HQ-trained \texttt{open-sci-ref} model is competitive with, and in
several languages stronger than, EuroLLM-1.7B on translated MMLU variants.
Figure~\ref{fig:belebele} shows a similar pattern during training: although
Nemotron-CC-HQ is treated as English-dominant in the main experiments, several
non-English Belebele languages rise above the 25\% random baseline late in
training.

We treat this result as diagnostic rather than conclusive. It may reflect
transfer from strong English pretraining, properties of translated benchmark
formats, or non-English content admitted by the language filtering threshold
used in Nemotron-CC-HQ. Distinguishing these explanations requires dataset
composition audits and controlled multilingual ablations. The result therefore
reinforces the broader reproducibility point: benchmark behavior should be
interpreted together with transparent information about data composition and
filtering.

\section{Limitations}

\texttt{open-sci-ref-0.02} is intended as a controlled reference suite, not an exhaustive map of the language model design space. The present release focuses on dense decoder-only transformers up to 1.7B parameters, trained with one tokenizer, one architecture family, and one tuned optimization procedure. This control is useful for dataset and compute comparisons, but it also means that the results may not transfer unchanged to other architectures, tokenizers, optimizers, or mixture-of-experts models.

Our scaling results should be read as reference trajectories rather than full scaling laws. We compare selected model sizes and token budgets on a common compute axis, which is enough to identify strong and weak procedures relative to shared baselines, but not enough to define compute-optimal frontiers or extrapolate precisely to much larger scales. Full scaling law estimation would require denser measurements and additional hyperparameter tuning.

The evaluation suite is also limited. The 11-task average provides a useful standardized summary, but it is mostly English-centered and language understanding based, and does not fully cover open-ended generation, long-context use, code, mathematical reasoning, factuality, safety, or post-training behavior. Task-level results show that dataset rankings can vary by benchmark, so small average score differences should not be overinterpreted.

Finally, comparisons to external models and multilingual benchmarks should be treated as diagnostic. External models differ from \texttt{open-sci-ref} in tokenizer, architecture details, data mixture, filtering, language allocation, and training recipe. For the multilingual benchmarks, the observed non-English performance may reflect transfer from English pretraining, translated benchmark artifacts, or
non-English content admitted by filtering thresholds.

\section{Conclusion \& Outlook}\label{sec:conclusion}

We introduced \texttt{open-sci-ref-0.02}, a suite of open dense transformer reference baselines for reproducible language model and dataset comparison. The suite spans four model scales, from 0.13B to 1.7B parameters, nine reference datasets, and token budgets up to 1T. By holding the architecture, tokenizer, optimization procedure, evaluation protocol, and compute accounting fixed across runs, and by releasing model weights, intermediate checkpoints, logs, code, and downstream evaluations, \texttt{open-sci-ref} provides a reference against which new datasets, model variants, and training recipes can be fairly compared.

The established references make dataset and training procedure comparisons more interpretable. Across the conventional reference datasets, rankings are broadly stable across model scales and training duration, with larger models making dataset differences more visible. The 1.7B model trained on 1T Nemotron-CC-HQ tokens provides the strongest reference point in this release, matching or slightly exceeding the average downstream performance of SmolLM2-1.7B while using roughly 11$\times$ less estimated training compute. These results make Nemotron-CC-HQ a useful high-performing anchor for future comparisons under the released recipe.

This stability, however, is not universal. MixtureVitae exhibits the largest upward rank change, moving from eighth of nine datasets at 0.13B to fourth at 1.7B. A single small-scale run would therefore have understated its relative performance at the largest scale evaluated here. This is the case that motivates multi-scale reference trajectories rather than single proxy runs: ranks measured at a single reference scale can strongly mislead about ranking at larger, unseen scales.

The same reference suite also clarifies comparisons to external models. Placing models on a common compute axis shows when parameter count or final score comparisons obscure large differences in training budget. In particular, the EuroLLM-1.7B comparison illustrates how a model can appear competitive by scale or by a steep trend to a larger model, while still underperforming transparent reference points trained with substantially less compute. 

Finally, the extended dataset set supports more controlled comparisons of permissively licensed and mixed-license corpora. At the 1.7B/300B scale, CommonCorpus-en and Comma-0.1 have a large gap to the non-permissive references, while MixtureVitae comes within 0.01 of the strong non-permissive baseline FineWeb-Edu on the 11-task average, driven by MMLU and CommonSenseQA (Sec.~\ref{sec:permissive-comparison}). This result suggests that also permissive dataset mixtures can be performant, if mixture ratio design, filtering, and the front-loading of reasoning and instruction data are taken care of. 

Looking forward, natural extensions include scaling the permissive reference datasets to larger token budgets, evaluating mixture-of-experts architectures under the same protocol, adding broader evaluation coverage, and deriving full scaling laws from denser measurements. More broadly, \texttt{open-sci-ref} is intended as public experimental infrastructure: a reusable set of references that researchers can inspect, reproduce and extend when evaluating new datasets, models, and training procedures.

\section*{Acknowledgements}
\label{sec:acknowledgements}

MN, JF, TC, NA, VK and JJ acknowledge co-funding by EU from Digital Europe Programme under grant no. 101195233 (openEuroLLM) and funding by EU Horizon under grant no. 101214398 (ELLIOT). MN and JJ acknowledge co-funding from EuroHPC Joint Undertaking programme under grant no. 101182737 (MINERVA), as well as funding by the German Federal Ministry of Research, Technology and Space (BMFTR) under grant no. 01IS24085C (OPENHAFM), under the grant 16HPC117K (MINERVA) and under the grant no. 01IS22094B (WestAI - AI Service Center West).

We gratefully acknowledge the Gauss Centre for Supercomputing e.V. for funding this work by providing computing time through the John von Neumann Institute for Computing (NIC) on the supercomputer JUWELS Booster at Jülich Supercomputing Centre (JSC), EuroHPC Joint Undertaking for computing time and storage on the EuroHPC supercomputer LEONARDO, hosted by CINECA (Italy) and the LEONARDO consortium through an EuroHPC Extreme Access grant EHPC-EXT-2023E02-068, storage resources on JUST granted and operated by JSC and supported by Helmholtz Data Federation (HDF), computing time granted by the JARA and JSC on the supercomputer JURECA at JSC, and computing time granted on prototype JEDI via JUREAP (JUPITER Early Access Program) grant at JSC.

Further thanks go to support provided by supercomputing facilities and their teams, especially to Damian Alvarez and Mathis Bode from Juelich Supercomputer Center (JSC, Germany) and to Laura Morselli and Domitilla Brandoni from CINECA (Italy).

We thank further for support \& advising: Sampo Pyysalo, Guilherme Penedo, Quentin Anthony, Hynek Kydlí\v{c}ek, Mehdi Cherti, Tomer Porian, Adam Ibrahim, Jonathan Burdge, Aaron Klein, Ken Tsui and Huu Nguyen.

We also would like to express gratitude to all the people who are working on making code, models and data publicly available, advancing community based research and making research more reproducible. Specifically, we would like to thank Open-$\Psi$ (Open-Sci) Collective\footnote{\url{https://discord.gg/GsKh4mBVcv}}, openEuroLLM Consortium\footnote{\url{https://openeurollm.eu}} and all the members of the LAION Discord server\footnote{\url{https://discord.gg/BZqhreFazY}} community for providing fruitful ground for scientific exchange and open-source development.

\bibliography{references,add_reference}

\appendix

\begin{center}
{\Large\bf Supplementary}
\end{center}

\section{Training Recipe and Hyperparameters}
\label{app:hparams}

This section gives the hyperparameter details for the training recipe used in the
\texttt{open-sci-ref-0.02} reference runs. We tune the recipe at the 1.3B scale
using 50B-token runs on C4 and FineWeb-Edu, selecting configurations by
validation loss. We first tune a standard cosine learning rate schedule
(Table~\ref{tab:cosine_hyperparameters}), then switch to a constant learning rate
schedule with linear cooldown, also known as WSD or trapezoid schedule
(Table~\ref{tab:wsd-hparams}). The WSD schedule is used for the released
reference runs because it matches or improves on the best cosine validation
loss while making it easier to resume training from intermediate checkpoints
and to extend runs to longer token budgets.

Unless otherwise stated, the reference runs use weight decay 0.05, a global
batch size of approximately 4M tokens, and a cooldown lasting 20\% of the
total training duration. Warmup length and learning rate depend on the target
token budget. For token budgets of 300B and above, we use a longer warmup of
25,000 iterations, which improved performance in tuning runs.

\begin{table*}[hbp]
\centering
\small
\begin{tabular}{lrrrr}
\hline
Token budget & Global batch & Iterations & Learning rate & Warmup \\
\hline
50B  & 2.65M & 18,839  & $3 \times 10^{-3}$ & 6,000 \\
50B  & 4.03M & 12,406  & $6 \times 10^{-3}$ & 1,000 \\
300B & 2.09M & 143,052 & $3 \times 10^{-3}$ & 5,000 \\
\hline
\end{tabular}
\caption{
Cosine learning rate configurations used during recipe tuning. These runs were
used to establish validation loss targets before switching to the WSD schedule.
}
\label{tab:cosine_hyperparameters}
\end{table*}

\begin{table*}[hbp]
\centering
\small
\begin{tabular}{lrrrrr}
\hline
Token budget & Global batch & Iterations & Learning rate & Warmup & Cooldown \\
\hline
50B  & 2.65M & 18,839  & $2 \times 10^{-3}$ & 6,000  & 3,767  \\
50B  & 4.12M & 11,921  & $4 \times 10^{-3}$ & 1,000  & 2,384  \\
300B & 2.09M & 143,052 & $1 \times 10^{-3}$ & 5,000  & 28,610 \\
300B & 4.12M & 72,661  & $4 \times 10^{-3}$ & 25,000 & 14,532 \\
1T   & 4.12M & 242,204 & $4 \times 10^{-3}$ & 25,000 & 48,440 \\
\hline
\end{tabular}
\caption{
WSD learning rate configurations. Cooldown is 20\% of the total training
duration. The 4.12M-token global batch rows define the main reference recipe
used for the longer 300B- and 1T-token runs.
}
\label{tab:wsd-hparams}
\end{table*}

\section{Scope and interpretation of the controlled comparisons}
\label{app:scope-of-control}

The term \emph{controlled comparison} in this work refers to
matched contrasts among \texttt{open-sci-ref} runs. It does
not imply that the suite measures a dataset's protocol-independent
or best-achievable quality. For two runs with the same model size
and token budget, the relevant experimental intervention is the
replacement of the pretraining dataset while retaining the shared
tokenizer, architecture family, training procedure, and evaluation
setup. The resulting difference should therefore be interpreted as
the effect of substituting the dataset \emph{under the
\texttt{open-sci-ref} protocol}.

Holding a component fixed does not make it neutral.
For example, the shared GPT-NeoX-20B tokenizer may represent some domains, formats, or languages more efficiently than others.
Likewise, the common optimization, sampling, and sequence packing procedure may interact differently with different corpus
compositions and repetition patterns. Relative dataset performance could therefore change under another tokenizer, architecture
family, optimization procedure, or sampling policy.

The shared recipe was selected using 1.3B-parameter, 50B-token
experiments on C4 and FineWeb-Edu, with configurations selected by validation loss, and was subsequently applied to the remaining
datasets without dataset-specific retuning. This choice is intentional: independently optimizing the training recipe for each
dataset would confound the dataset comparison with differences in hyperparameter search effort and would make the resulting models
less useful as common reference points. The corresponding tradeoff is that the suite measures performance under one shared recipe,
rather than the best performance attainable from each dataset.
Because C4 and FineWeb-Edu informed recipe selection, the selected settings may be better matched to those corpora than to some other
mixtures. This release does not claim otherwise or perform a separate learning rate optimization for every dataset.

Not every numerical configuration is globally identical across the suite. Model dimensions are prescribed by model scale, while
learning rate, warmup, and cooldown settings are prescribed by the
target training configuration, as reported in
Tables~\ref{tab:model_architecture},
\ref{tab:cosine_hyperparameters}, and
\ref{tab:wsd-hparams}. These prescribed values are applied consistently across datasets within each matched comparison.
Consequently, direct dataset comparisons should be made within a common model size and token budget setting, while cross-scale results should be read as trajectories under the reference protocol.

\section{Additional experimental results}
\label{sec:additional_exp_results}

\subsection{Further reference baselines}
\label{subsec:further_reference_baselines}

\begin{figure}[htbp]
    \centering
    \includegraphics[width=\columnwidth]{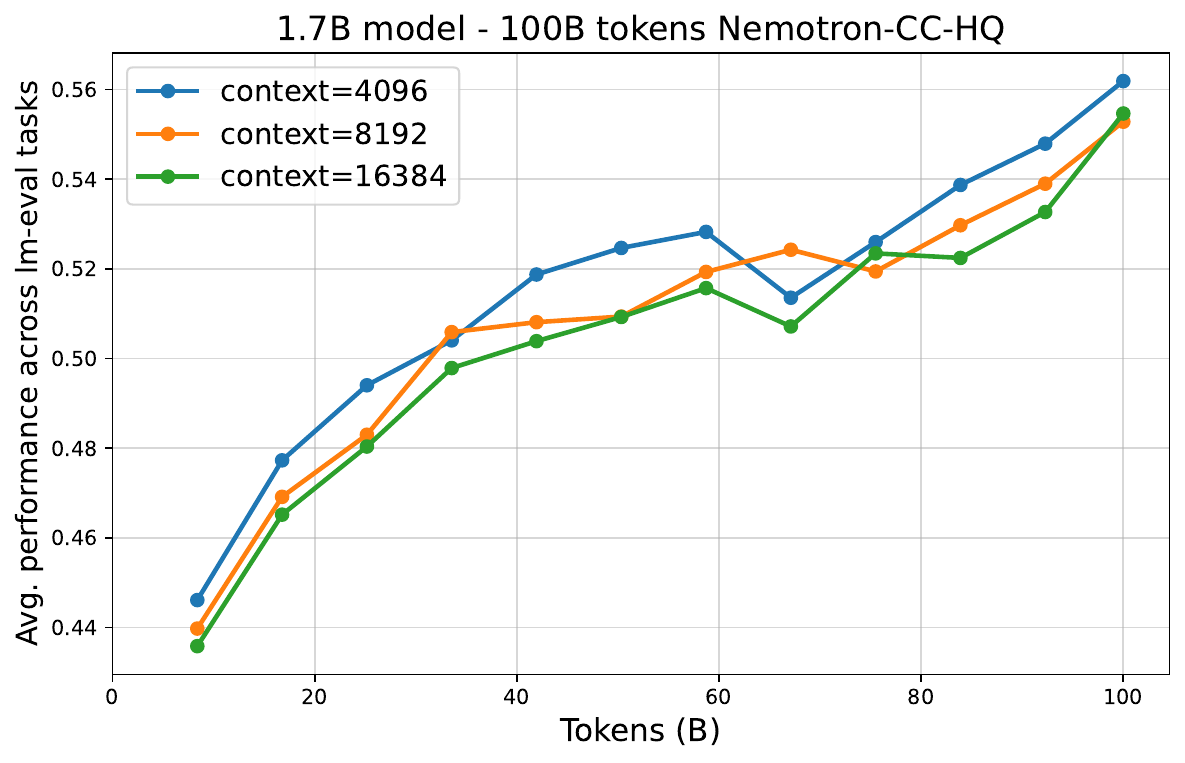}
    \caption{Comparing training with different context lengths. Training run for \texttt{\opensciref{}} 1.7B model on 100B tokens of Nemotron-CC-HQ, using various context lengths of 4096 (RoPE base 100k), 8192 (RoPE base 500k) and 16384 (RoPE base 1M). We observe no difference in average performance, ensuring models with larger context length have no drop in performance compared to reference context length of 4096.}
\label{fig:contextlength_reference}
\end{figure}

\textbf{Context length.} We also perform training with various context lengths while making sure that the global batch size remains constant across different variants. We train using context lengths of 4096, 8192, and 16384, and observe no difference in downstream evaluations (Suppl. Fig.~\ref{fig:contextlength_reference}). When varying context length, we also adapt the base for RoPE correspondingly (100k for 4096, 500k for 8192, and 1M for 16384).

\textbf{Dropout.} We also perform a test enabling or disabling dropout (hidden and attention), without observing any relevant difference in performance. The released reference models have dropout set to 0.

\subsection{Robustness of aggregate rankings}
\label{app:aggregate-rank-robustness}

The 11-task average provides a compact summary of the evaluation suite, but it
should not be interpreted as a calibrated, task-independent measure of overall
model quality. The component evaluations differ in their metrics, chance levels,
content, and discriminatory power. We therefore test whether the principal
ranking conclusions depend on model scale or on the inclusion of any one
evaluation.

\paragraph{Cross-scale rank stability.}
At each model scale, we rank datasets by their 300B-token endpoint average, with
rank 1 denoting the highest average, and compute Spearman rank correlations
between scales. Table~\ref{tab:scale-rank-robustness} reports two versions of
this analysis. The first covers the six established reference datasets:
Nemotron-CC-HQ, DCLM, HPLT-2.0, FineWeb-Edu, Pile, and SlimPajama. The second
covers the eight datasets with reported 300B-token endpoints at all four model
scales in Fig.~\ref{fig:reference300b}, additionally including MixtureVitae and
Comma-0.1. CommonCorpus-en is omitted from the second analysis because its 0.4B
endpoint is not among the reported trajectories.

\begin{center}
\begin{minipage}{\columnwidth}\centering
\small
\setlength{\tabcolsep}{5pt}
\begin{adjustbox}{max width=\columnwidth}
\begin{tabular}{lcc}
\toprule
\textbf{Scale pair} &
\textbf{Established references} &
\textbf{Complete trajectories} \\
&
\textbf{($n=6$)} &
\textbf{($n=8$)} \\
\midrule
0.13B--0.4B & 0.943 & 0.952 \\
0.13B--1.3B & 0.943 & 0.833 \\
0.13B--1.7B & 0.943 & 0.738 \\
0.4B--1.3B  & 0.886 & 0.881 \\
0.4B--1.7B  & 0.886 & 0.810 \\
1.3B--1.7B  & 1.000 & 0.976 \\
\midrule
Median       & 0.943 & 0.857 \\
\bottomrule
\end{tabular}
\end{adjustbox}
\captionof{table}{
Spearman correlations between dataset rankings at the 300B-token endpoint. The
established-reference analysis covers Nemotron-CC-HQ, DCLM, HPLT-2.0,
FineWeb-Edu, Pile, and SlimPajama. The complete-trajectory analysis adds
MixtureVitae and Comma-0.1, the two additional datasets with reported endpoints
at all four scales. Higher values indicate more stable ordering across model
scales.}
\label{tab:scale-rank-robustness}
\end{minipage}
\end{center}

Among the six established references, pairwise rank correlations range from
0.886 to 1.000, with a median of 0.943. Their ordering is therefore largely
preserved across scale, and is identical between the 1.3B and 1.7B endpoints.
The broader complete-trajectory analysis has correlations between 0.738 and
0.976, with a median of 0.857. The reduction is primarily associated with
MixtureVitae, which moves from rank 8 at 0.13B to rank 7 at 0.4B, rank 5 at
1.3B, and rank 4 at 1.7B. Thus, the observed stability is strong for the
established reference datasets but is not universal, consistent with the
scale-dependent MixtureVitae result discussed in the main text.

\paragraph{Leave-one-task-out sensitivity.}
We next test whether the 1.7B/300B ranking is driven by a single evaluation. For
each task $j$, we omit that task and recompute the average over the remaining
ten evaluations:
\begin{equation}
    \bar{s}^{(-j)}_m =
    \frac{1}{10}\sum_{k \neq j} s_{m,k},
\end{equation}
where $s_{m,k}$ is the score of model $m$ on evaluation $k$. We then rerank all
nine datasets separately for each of the eleven omissions.
Table~\ref{tab:loto-rank-robustness} reports each dataset's rank under the full
11-task average and the minimum and maximum ranks it attains across the
leave-one-task-out averages.

\begin{center}
\begin{minipage}{\columnwidth}\centering
\small
\setlength{\tabcolsep}{5pt}
\begin{adjustbox}{max width=\columnwidth}
\begin{tabular}{lccc}
\toprule
\textbf{Dataset} &
\textbf{11-task avg.} &
\textbf{Full rank} &
\textbf{LOTO rank range} \\
\midrule
Nemotron-CC-HQ   & 0.64 & 1 & 1--1 \\
DCLM             & 0.59 & 2 & 2--2 \\
FineWeb-Edu      & 0.57 & 3 & 3--4 \\
MixtureVitae     & 0.56 & 4 & 3--7 \\
HPLT-2.0         & 0.56 & 5 & 4--5 \\
SlimPajama       & 0.55 & 6 & 5--6 \\
Pile             & 0.54 & 7 & 6--7 \\
Comma-0.1        & 0.50 & 8 & 8--8 \\
CommonCorpus-en  & 0.47 & 9 & 9--9 \\
\bottomrule
\end{tabular}
\end{adjustbox}
\captionof{table}{
Leave-one-task-out robustness of the dataset ranking at the 1.7B-parameter,
300B-token endpoint. For each evaluation, we recompute the average over the
other ten evaluations and rerank all nine datasets. Averages are displayed to
two decimal places, while ranks are calculated from the unrounded endpoint
scores. Rank 1 is best.}
\label{tab:loto-rank-robustness}
\end{minipage}
\end{center}

The leave-one-task-out ranking remains close to the full ranking: its Spearman
correlation with the full ranking ranges from 0.900 to 1.000 across the eleven
omissions, with a median of 1.000. Omitting LAMBADA, OpenBookQA, WinoGrande,
ARC-Challenge, ARC-Easy, or PIQA leaves the entire ranking unchanged. Omitting
MMLU or BoolQ exchanges the positions of MixtureVitae and HPLT-2.0, while
omitting COPA or HellaSwag exchanges the positions of MixtureVitae and
FineWeb-Edu.

CommonSenseQA has the largest effect on the middle of the ranking. When it is
omitted, MixtureVitae moves from rank 4 to rank 7, while HPLT-2.0, SlimPajama,
and Pile each move up by one position. Conversely, when COPA or HellaSwag is
omitted, MixtureVitae moves to rank 3 and FineWeb-Edu to rank 4. These results
support the stable placement of Nemotron-CC-HQ and DCLM at the top and
Comma-0.1 and CommonCorpus-en at the bottom. They also show that the small
average differences among FineWeb-Edu, MixtureVitae, and HPLT-2.0 should be
interpreted as a task-composition-dependent middle group rather than as a strict
and universal ordering.

These analyses measure sensitivity to scale and to the composition of the
evaluation suite, conditional on the realized trained checkpoints. They do not
quantify uncertainty from finite evaluation samples or variation across training
seeds.

\section{Distributed training on supercomputers}
\label{sec:supercomputers}
We make use of various publicly funded European supercomputers to perform distributed training necessary for obtaining reference models. We conducted training experiments using Megatron-LM, employing NVIDIA containers for each machine. We used Leonardo (CINECA, 64GB A100), JUWELS Booster (JSC, 40GB A100), JEDI (JSC, early JUPITER prototype, 96GB H100), and JUPITER (JSC, 96GB H100). For each model scale, a combination of GPU numbers was selected that was providing good throughput and GPU utilization given the global batch size. We obtain good GPU utilization of 157--201 TFLOP/s/GPU on A100 machines and 320--395 TFLOP/s/GPU on H100 machines while working with hundreds of GPUs. The smallest 0.13B model is the exception at 114 TFLOP/s/GPU. Tab.~\ref{tab:leonardo_training} reports the per-run configurations and measured throughput on Leonardo, and Tab.~\ref{tab:open_sci_ref_runs} the corresponding runs on JEDI, JUPITER and JUWELS Booster.

\begin{table*}[htbp]
\centering
\begin{adjustbox}{width=\textwidth}
\begin{tabular}{c r c r l c c l l}
\hline
\shortstack{Params\\(B)} & GPUs & \shortstack{Micro\\batch} & \shortstack{Context\\length} & \shortstack{Global batch\\(samples/tokens)} & \shortstack{TFLOP/s/\\GPU} & \shortstack{Tokens/s/\\GPU} & \shortstack{Run time, h\\(50B/300B/1T)} & \shortstack{A100 GPU-h\\(50B/300B/1T)} \\
\hline
0.13 & 84  & 12 & 4096  & 1008/4.12M & 114 & 87710 & 1.89/11.31/37.70  & 158/950/3167 \\
0.4  & 100 & 10 & 4096  & 1000/4.09M & 157 & 44550 & 3.12/18.71/62.35  & 312/1871/6235 \\
1.3  & 252 & 4  & 4096  & 1008/4.12M & 172 & 16800 & 3.28/19.68/65.60  & 827/4959/16530 \\
1.3  & 216 & 6  & 2048  & 1296/2.65M & 167 & 18490 & 3.48/20.87/69.56  & 751/4508/15026 \\
1.7  & 252 & 4  & 4096  & 1008/4.12M & 184 & 14490 & 3.80/22.81/76.05  & 958/5749/19164 \\
1.7  & 252 & 2  & 8192  & 504/4.12M  & 186 & 12320 & 4.47/26.84/89.47  & 1127/6764/22546 \\
1.7  & 252 & 1  & 16384 & 252/4.12M  & 201 & 10080 & 5.47/32.80/109.32 & 1377/8265/27549 \\
\hline
\end{tabular}
\end{adjustbox}
\caption{\texttt{open-sci-ref} reference baseline distributed training runs on
Leonardo. The TFLOP/s/GPU column reports the throughput accounting used during
training, which adds the attention term $12\,L\,s\,h$ (layers, sequence length,
hidden size) to the $6N$ per-token cost. Tab.~\ref{tab:model_architecture} and
the compute values in Tab.~\ref{tab:standard-endpoint-results} use the $6N$
estimate alone, so the two conventions differ by that term.}
\label{tab:leonardo_training}
\end{table*}

\begin{table*}[htbp]
\centering
\footnotesize
\begin{subtable}{\textwidth}
\centering
\begin{tabular}{c c c c l c c l l}
\hline
\shortstack{Params\\(B)} & GPUs & \shortstack{Micro\\batch} & \shortstack{Context\\length} & \shortstack{Global batch\\(samples/tokens)} & \shortstack{TFLOP/s/\\GPU} & \shortstack{Tokens/s/\\GPU} & \shortstack{Run time, h\\(50B/300B/1T)} & \shortstack{H100 GPU-h\\(50B/300B/1T)} \\
\hline
1.3 & 64 & 16 & 2048 & 1024/2.09M & 320 & 35370 & 6.13/36.81/122.70 & 393/2356/7853 \\
1.7 & 128 & 2 & 16384 & 256/4.19M & 395 & 19820 & 5.48/32.85/109.50 & 701/4205/14016 \\
\hline
\end{tabular}
\caption{executed on JEDI}
\end{subtable}

\begin{subtable}{\textwidth}
\centering
\begin{tabular}{c c c c l c c l l}
\hline
\shortstack{Params\\(B)} & GPUs & \shortstack{Micro\\batch} & \shortstack{Context\\length} & \shortstack{Global batch\\(samples/tokens)} & \shortstack{TFLOP/s/\\GPU} & \shortstack{Tokens/s/\\GPU} & \shortstack{Run time, h\\(50B/300B/1T)} & \shortstack{H100 GPU-h\\(50B/300B/1T)} \\
\hline
1.7 & 128 & 8 & 4096 & 1024/4.19M & 370 & 29170 & 3.72/22.32/74.39 & 476/2856/9521 \\
\hline
\end{tabular}
\caption{executed on JUPITER}
\end{subtable}

\begin{subtable}{\textwidth}
\centering
\begin{tabular}{c c c c l c c l l}
\hline
\shortstack{Params\\(B)} & GPUs & \shortstack{Micro\\batch} & \shortstack{Context\\length} & \shortstack{Global batch\\(samples/tokens)} & \shortstack{TFLOP/s/\\GPU} & \shortstack{Tokens/s/\\GPU} & \shortstack{Run time, h\\(50B/300B/1T)} & \shortstack{A100 GPU-h\\(50B/300B/1T)} \\
\hline
1.3 & 328 & 6 & 2048 & 1968/4.03M & 158 & 17400 & 2.43/14.60/48.66 & 798/4788/15961 \\
\hline
\end{tabular}
\caption{executed on JUWELS Booster}
\end{subtable}
\caption{\texttt{open-sci-ref} reference baseline distributed training runs. The
TFLOP/s/GPU column follows the same accounting as Tab.~\ref{tab:leonardo_training}.}
\label{tab:open_sci_ref_runs}
\end{table*}

\end{document}